\documentclass[letterpaper]{article} % DO NOT CHANGE THIS
\usepackage{aaai25}  % DO NOT CHANGE THIS
\usepackage{times}  % DO NOT CHANGE THIS
\usepackage{helvet}  % DO NOT CHANGE THIS
\usepackage{courier}  % DO NOT CHANGE THIS
\usepackage[hyphens]{url}  % DO NOT CHANGE THIS
\usepackage{graphicx} % DO NOT CHANGE THIS
\usepackage{natbib}  % DO NOT CHANGE THIS AND DO NOT ADD ANY OPTIONS TO IT
\usepackage{caption} % DO NOT CHANGE THIS AND DO NOT ADD ANY OPTIONS TO IT
\usepackage{algorithm}
\usepackage{algorithmic}
\usepackage{tabularx}
\usepackage{multirow}
\usepackage{multicol}
\usepackage{booktabs}
\usepackage{color}
\usepackage{subfig}
\usepackage{graphicx}
\usepackage{caption}
\usepackage{amsmath}
\usepackage{amssymb}
\usepackage{makecell}

\usepackage[english]{babel}

\usepackage{cleveref}
\crefname{figure}{Fig.}{Figs.}
\crefname{table}{Table.}{Tables.}
\crefname{section}{Section}{Secs.}

\usepackage{xcolor}

\usepackage{newfloat}
\usepackage{listings}
\DeclareCaptionStyle{ruled}{labelfont=normalfont,labelsep=colon,strut=off} % DO NOT CHANGE THIS
\floatstyle{ruled}
\newfloat{listing}{tb}{lst}{}
\floatname{listing}{Listing}
\title{L3TC: Leveraging RWKV for Learned Lossless Low-Complexity Text Compression}
\author {
    Junxuan Zhang\textsuperscript{\rm 2}\equalcontrib,
    Zhengxue Cheng\textsuperscript{\rm 1}\equalcontrib\thanks{Corresponding author.},
    Yan Zhao\textsuperscript{\rm 1},
    Shihao Wang\textsuperscript{\rm 2},\\
    Dajiang Zhou\textsuperscript{\rm 2},
    Guo Lu\textsuperscript{\rm 1},
    Li Song\textsuperscript{\rm 1}
}
\affiliations {
    \textsuperscript{\rm 1}Institute of Image Communication and Network Engineering, Shanghai Jiao Tong University\\
    \textsuperscript{\rm 2}Ant Group, Hangzhou, China\\
    junxuan.zjx@antgroup.com, zxcheng@sjtu.edu.cn, zhaoyanzy@sjtu.edu.cn, shihao.wsh@antgroup.com, dajiang.zdj@antgroup.com, luguo2014@sjtu.edu.cn, song\_li@sjtu.edu.cn
}

\usepackage{bibentry}
\usepackage[utf8]{inputenc}
\DeclareUnicodeCharacter{FF0C}{,}

\begin{document}

\maketitle

\begin{abstract}

% Large language models (LLMs) exhibit powerful predictive capabilities and can be used as probabilistic models for data compression. 
Learning-based probabilistic models can be combined with an entropy coder for data compression. However, due to the high complexity of learning-based models, their practical application as text compressors has been largely overlooked. To address this issue, our work focuses on a low-complexity design while maintaining compression performance. We introduce a novel Learned Lossless Low-complexity Text Compression method (L3TC). Specifically, we conduct extensive experiments demonstrating that RWKV models achieve the fastest decoding speed with a moderate compression ratio, making it the most suitable backbone for our method. Second, we propose an outlier-aware tokenizer that uses a limited vocabulary to cover frequent tokens while allowing outliers to bypass the prediction and encoding. Third, we propose a novel high-rank reparameterization strategy that enhances the learning capability during training without increasing complexity during inference. Experimental results validate that our method achieves 48\% bit saving compared to gzip compressor. Besides, \emph{L3TC} offers compression performance comparable to other learned compressors, with a $50\times$ reduction in model parameters. More importantly, \emph{L3TC} is the fastest among all learned compressors, providing real-time decoding speeds up to megabytes per second. Our code is available at \color{gray}{\textit{https://github.com/alipay/L3TC-leveraging-rwkv-for-learned-lossless-low-complexity-text-compression.git}}.

\end{abstract}


\section{Introduction}

Lossless text compression is a fundamental research field focused on reducing data size based on information theory. In 1948, Shannon~\cite{shannon:1948} described that maximizing the $\text{log}_2$-likelihood of data is equivalent to minimizing the number of bits required for compression. Then information theory~\cite{MacKay:2003} established the essential equivalence between probabilistic models of data and lossless compression. Recently, learning-based probabilistic models, like language models, have developed rapidly and demonstrated remarkable in-context learning capabilities across various tasks by predicting the probability of the next token. This predictive capability can be inherently combined with entropy encoding to serve as a data compressor. Entropy encoding can be implemented in various ways, including Huffman Coding~\cite{Huffman:1952}, Arithmetic Coding (AC)~\cite{AC:1977, AC:1991}, and Asymmetric Numeral Systems (ANS)~\cite{ANS:2013}.

% \begin{figure}[ht]
% \centering
% \includegraphics[width=0.95\columnwidth]{Figure/teaser.pdf} 
% % \vspace{-3mm}
% \captionsetup{font=small}
% \caption{Compression Ratio (CR) vs. Model Size: Notable compressors, including gzip, and popular learning-based models are compared. \emph{L3TC} reports $50\times$ model size reductions with comparable compression performance. Meanwhile, although other learned models typically decode at KB/s speeds, \emph{L3TC} achieves decoding speeds up to MB/s.}
% \vspace{-3mm}
% \label{fig1:teaser}
% \end{figure}

\begin{figure}[t]
\centering
\includegraphics[width=0.88\columnwidth]{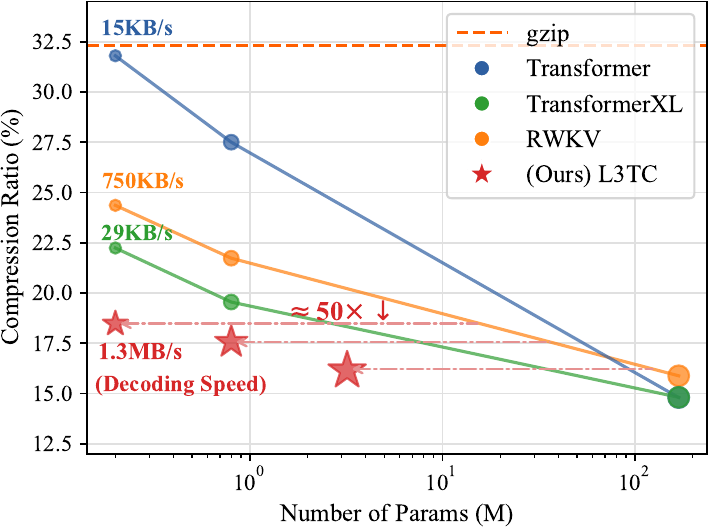} 
\vspace{-2mm}
\captionsetup{font=small}
\caption{\small Compression Ratio vs. Model Size: Notable compressors, including gzip and learning-based models are compared. \emph{L3TC} achieves the best compression ratio among them and reports $50\times$ model size reductions with comparable compression performance. When running on devices, other learned models typically decode at KB/s speeds, while \emph{L3TC} achieves decoding speeds up to MB/s.}
\vspace{-2mm}
\label{fig1:teaser}
\end{figure}

% Compression Ratio (CR) versus Model size. Notable compressors, like gzip, pretrained LLMs and corresponding small models are compared. \emph{L3TC} reports a $50\times$ model size reduction with a comparable CR. Besides, pretrained Llama models usually decode at speeds in B/s, while \emph{L3TC} achieves decoding speeds up to MB/s.

% Raw compression ratio with different models.
% \textbf{Second Paragraph}

% TODO: propose 中间那个大长句 ("because one-token ... the inference complexity") 改简单一点 (去掉 one-token 的逻辑), 比如 since the decoding complexity is nearly equivalent to the model's inference complexity. -> 现在定语太多了比较绕.
Despite the success of language models in many intelligent tasks, the literature has largely overlooked their application as a practical lossless text compressor. This is not surprising, because the decoding complexity is nearly equivalent to the model's inference complexity, which is significantly slower compared to popular engineered compressors such as gzip~\cite{gzip}. For a practical usage, the data compression task basically demands a very critical real-time decoding performance, up to megabytes per second. Therefore, it is a very promising and challenging research direction to develop a low-complexity text compression algorithm based on the learning-based likelihood prediction.
% TODO: propose 不提 although already lightweight LLMs, 直接说 develop low-complexity compressor. -> 感觉说其他轻量化方法不够实时有可能不太好 是吗
% propose 讲一下 complexity 和 efficiency 的 trade-off, 是 L3TC 区别于仅 light-weight / 仅 efficient 的地方.

% \textbf{Third Paragraph}
Several existing works have demonstrated the effectiveness of learned probabilistic models based on transformers and LSTMs for lossless text compression. Specifically, nncp v3.2~\cite{nncpv2:2021}, which is based on transformers, and cmix v20~\cite{cmix:2023}, which utilizes LSTMs, achieve the top-2 compression efficiencies on the enwiki9 dataset~\cite{benchmark}. However, they both involve significant decoding complexity, taking 2.8 days and 7.2 days to decode 1GB of text, respectively. Recently, LLMZip~\cite{llmzip:2023} demonstrates that using LLama-7B~\cite{touvron2023llama} with Arithmetic Coding can achieve about 40\% bit rate savings compared to PAQ8H~\cite{paq8h} on text8 dataset~\cite{text-data}, but the complexity is extremely high. Additionally, Google~\cite{LMisComp:2024} shows that Chinchilla-70B~\cite{LMisComp:2024} achieves more promising bit rate savings compared to gzip on enwiki9 dataset~\cite{hutterprize}. However, these models' sizes far exceed the data to be compressed, making them impractical for the real-world use.

% However, these models' sizes are much larger than the data size to be compressed, rendering it impractical.

% \textbf{Fourth Paragraph}
To address the aforementioned high-complexity limitations, we explore a Learned Lossless Low-Complexity Text Compression (\emph{L3TC}) method. It leverages the probability prediction of RWKV models while enhancing compression performance with two proposed novel designs. 1) We extensively evaluate the compression capability of various architectures, including Transformer, Transformer-XL, and RWKV, finding that RWKV offers the fastest decoding speed with moderate compression ratios. 2) We propose an outlier-aware tokenizer, focuses on frequent tokens while allowing unknown tokens to bypass encoding. 3) We introduce a novel high-rank reparameterization strategy to improve the learning capability during training without increasing inference complexity. Experiments show that \emph{L3TC} achieves a significant 48\% bit saving compared to gzip, and offers the fastest decoding speed among all the learned compressors.
% Experiments demonstrate the superior performance of \emph{L3TC}, by achieving 48\% bit saving than gzip. Besides, \emph{L3TC} is the fastest among learning-based compressors.

%Compared to widely-used \emph{gzip} compressor, our method can achieve 50\% bitrate saving. Benefiting our proposed outlier-aware tokenizer and high-rank reparameterization, out methods L3TC can achieve comparable compression ratio, with a over  model size reduction. Meanwhile, it is worthy noting that our method \emph{L3TC} is the fastest among learning-based compressors, achieving the decoding speed up up to MB/s.

% {\color{blue} We conduct extensive experiments on various LLM architectures and demonstrate that RWKV achieves the fastest inference speed with moderate compression ratio, making it suitable for practical use.} 
% TODO[Done]: propose 修改第一个 contrubution, 个人感觉现在的版本有点像陈述 RWKV 的优点 -> 要不讲一下 RWKV 作为我们的 backbone (题目"leverage RWKV") 变为比如说: We extensively evaluate various LLM architectures and select RWKV as our backbone due to its ability to provide satisfactory compression ratios with the fastest speed. 
The contributions of our work are summarized as follows:
\begin{itemize}
\item  We extensively evaluate various architectures and select RWKV as our backbone, owing to its acceptable compression ratios with the fastest speed.
\item We propose an outlier-aware tokenizer and a high-rank reparameterization strategy to improve the compression performance while maintaining a low-complexity design.
\item Experimental results validate that \emph{L3TC} achieves a 48\% bit saving compared to gzip, reduces the model size by over $50\times$ compared to other learned compressors with similar compression ratios, and offers decoding speeds up to the range of MB/s on mobile devices.
\end{itemize}

% and achieves over $50\times$ model size reduction compared to other learned compressors with comparable compression ratio. Meanwhile, \emph{L3TC} provides a decoding speed up to MB/s on devices.

% We propose two novel methods, outlier-aware tokenizer and high-rank reparameterization, to improve the compression performance with a low-complexity design.

% {\color{red}[Expect to reach the left half of the 2-nd page]}

% \vspace{-3mm}
\section{Related Work}

We review related work in three areas: classical text compressors, learned text compressors, and recent language models and their relation to compression.
% We review related works from three aspects, including classical and learned text compressors and the relation of large language models with compression.

\vspace{-1mm}
\paragraph{Classical Text Compressors.}  The development of lossless text compression has a long history. Typical compression tools include gzip, bzip2, and zstd. gzip~\cite{gzip} is based on the Deflate algorithm, which is a combination of LZ77~\cite{lz77:1977} and Huffman coding~\cite{Huffman:1952}. It works by searching for duplicate strings within a sliding window and replacing them with a pointer to their previous occurrence and the string length. BZIP2~\cite{bzip2} is another popular lossless text compressor that uses Burrows-Wheeler Transform and run-length encoding, Huffman Coding. Owing to its transform to rearrange the input text into runs of similar characters, bzip2 achieves a higher compression ratio than gzip. zstd~\cite{zstd}, a more recent algorithm, introduces several advanced techniques such as optimal parsing and dictionary compression, significantly improving the compression performance. 

% The basic idea is to search for duplicate strings within a sliding window of a certain size. When a duplicate string is found, it is replaced with a pointer to the previous occurrence of the string and the length of the string.

% From the viewpoint of entropy encoding, Huffman coding is widely used in combination with other compression algorithms, such as in the Deflate algorithm used in ZIP files. 

\vspace{-1mm}
\paragraph{Learned Text Compressors.} Transformer~\cite{transformer:2017} or LSTM~\cite{lstm:1997} based compression with arithmetic coding has achieved state-of-the-art compression performance according to the Large Text Compression Benchmark~\cite{benchmark} and Hutter Prize. Notable examples include cmix v20 and nncp. Cmix combines multiple modeling techniques, like context mixing, prediction by partial matching, and neural network-based models, to enhance compression. Nncp uses an online compression approach, where a pseudo-randomly initialized model is continuously trained on the data stream, adapting its weights to enhance predictions over time. Language models, with their strong in-context learning abilities, are well-suited for offline compression. For instance, LLMZip, an offline compression method using pretrained LLaMa-7B, achieves up to 0.71~\emph{bpc} (bit per character) on the text8 dataset, outperforming popular archive tools like PAQ8H (1.2bpc). However, these learned compressors are highly resource-intensive, requiring several days or a week to compress 1GB of text.

\begin{figure*}[ht]
\centering
\includegraphics[width=0.97\textwidth]{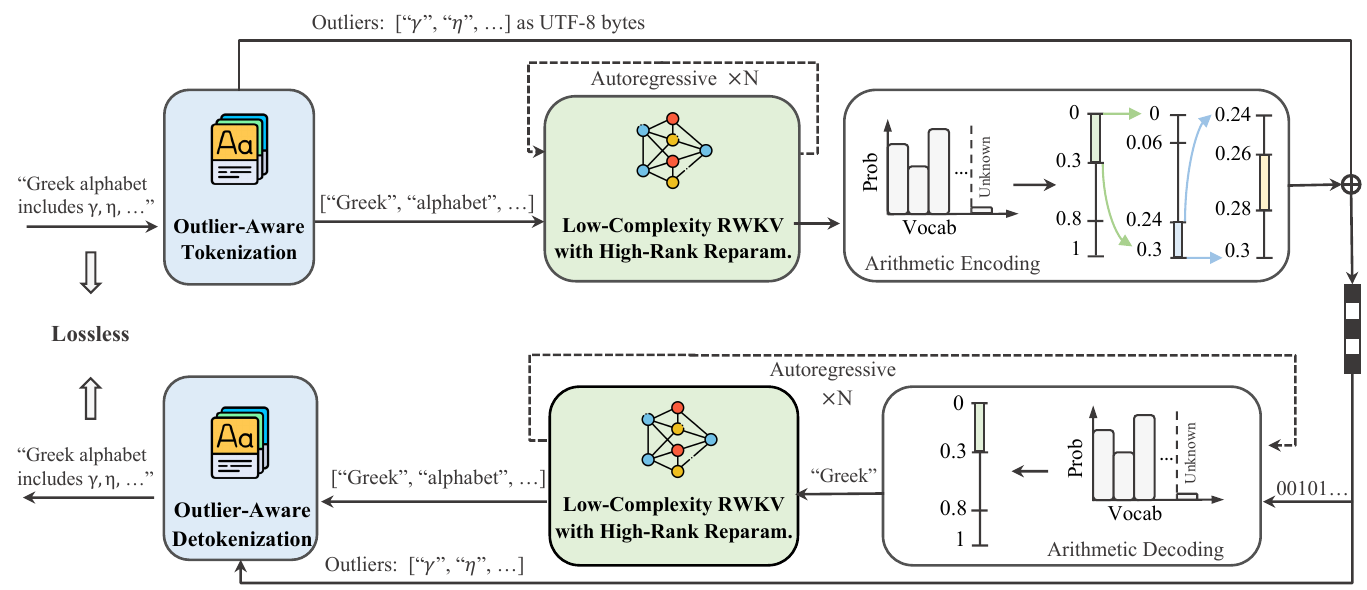} %
\vspace{-2mm}
\captionsetup{font=small}
\caption{\small Overall architecture of our proposed Learned Lossless Low-Complexity Text Compression (\emph{L3TC}), which consists of three primary components: tokenization, prediction and encoding. The text is firstly segmented through a novel outlier-aware tokenizer. Tokens in the vocabulary are then predicted by a low-complexity RWKV model and subsequently encoded by an arithmetic coder. Outliers, which appear infrequently, are allowed to bypass the prediction and encoding. Additionally, a high-rank reparameterization strategy is introduced to enhance the RWKV models's prediction capability during training without increasing inference complexity.}
\vspace{-3mm}
\label{fig2:overall}
\end{figure*}

% In the online setting, a pseudo-randomly initialized model is directly trained on the stream of data that is to be compressed, while the offline setting, which we consider in our work, trains the model on an external dataset before employing it to compress a (potentially different) data stream. Consequently, offline compression is performed in-context, with a fixed set of model parameters. Transformers have demonstrated impressive in-context learning abilities (Laskin et al., 2023; Brown et al., 2020; Wei et al., 2022; Genewein et al., 2023) and are thus ideally suited for offline compression.
\vspace{-1mm}
\paragraph{Language Models and Compression.} Recent studies have discussed the connection between compression and likelihood maximization in language models. \cite{huang2024} examines the linearity relationship between compression and intelligence for various tasks. Google~\cite{LMisComp:2024} demonstrates that Chinchilla-70B, pretrained on text datasets, not only achieves a promising compression ratio on text compression, but also compresses ImageNet~\cite{imagenet} patches to 48\% and LibriSpeech~\cite{librispeech} audios to 21\% of their raw size, obviously outperforming the domain-specific compressors like PNG (60\%)~\cite{png} and FLAC (30\%)~\cite{flac}. Additionally, various studies have investigated the relationship between tokenization and compression. \cite{trainingllmsneurallycompressed} introduces a novel text segmentation technique that significantly outperforms byte-level segmentation in terms of perplexity and inference speed. Pathpiece~\cite{tokenizationcompression} argues that fewer tokens do not always result in better downstream performance. \cite{unpackingtokenizationevaluatingtext} suggests that compression is a reliable intrinsic indicator for tokenization quality.

% Based on these works, promising improvements on compression ratio is observed. However, they do not address the significant complexity issue when models with billions of parameters are utilized. Instead, our work focuses on a low-complexity compression design to achieve a relatively high compression ratio. 

Though learning-based compression has shown promising improvements, the significant complexity of models with billions of parameters remains unaddressed. Therefore, our work prioritizes a low-complexity design while maintaining competitive compression ratios.

% Building on these works, promising improvements in compression ratios has been observed. However, they do not address the significant complexity involved when using models with billions of parameters. Instead, our work prioritizes a low-complexity compression design.

% We conduct an extensive investigation and explore the novel reparameterization strategy to achieve the best trade-off between compression ratio and inference speed.

% {\color{red} [Expect to reach the 1/4 of the 3-rd page]}

% TODO[Done]: 标题里有斜体看着有一丢丢怪, 并建议顺序变为 proposed L3TC method
% \section{{\color{blue} Proposed Method \emph{L3TC}}}

\section{Proposed Method}

\subsection{Overall Architecture}

The overall architecture of our proposed method \emph{L3TC} is shown in \cref{fig2:overall}. The text is firstly segmented into a sequence of tokens $x_{1:N}$, where $N$ denotes the total number of tokens. Then $x_i$ is fed into the models to calculate the probability based on the context information, i.e., $p(x_i | x_{\leq i})$. Arithmetic coding, known for achieving optimal coding efficiency, is used to compress the data to its entropy limit. The expected (optimal) bit number of the compressed data is the entropy:
\begin{equation}
    H(p) = \mathop{\mathbb{E}} (\sum_{i=1}^{n} -\text{log}_{2} p(x_i | x_{\leq i})])
    \label{eq.1}
\end{equation}

It can be found that language models are trained using cross-entropy loss to maximize the likelihood, which is consistent with compression objectives. At the receiver side, tokens are decoded autoregressively using the probabilistic model and an arithmetic decoder as \cref{fig2:overall}. Since the complexity of the arithmetic coder is usually smaller than that of model inference, the overall decoding complexity $\mathcal{O}(x_{1:N})$ is proportional to the number of tokens $N$ and the network inference time $\mathcal{O}(x_{i})$, i.e., 
\begin{equation}
 \mathcal{O}(x_{1:N}) =  \mathcal{O}(x_{i})\times N   
\end{equation}
To reduce the overall decoding complexity, we first explore a low-complexity RWKV model to minimize per-token inference time $\mathcal{O}(x_{i})$ in \cref{sec:3.2}. Then, we introduce an outlier-aware tokenizer that tokenizes only frequent characters and bypasses encoding for outliers, as detailed in \cref{sec:3.3}. Finally, we present a high-rank reparameterization strategy in \cref{sec:3.4} to enhance training without increasing inference complexity. All these innovations form our learned lossless low-complexity text compressor (\emph{L3TC}).

% We also propose an outlier-aware tokenizer to allowing outliers bypass the prediction and encoding, decreasing $N$ in Section 3.3.

\subsection{Low-Complexity RWKV Models} \label{sec:3.2}
% TODO: spam-char-128 with a coverage of 0.999 as the tokenizer -> 缩写有点不好懂，所以展开了一下: a character-based SentencePiece ~\cite{spm} tokenizer (SPM-CHAR) with coverage of 0.999 and vocabulary size of 128

% Context length is a key factor to compression ratio. In practice, to limit the complexity, we chunk the data into sequences of $C$ bytes, with $C$ set to 2048, following the settings in ~\cite{LMisComp:2024}.

% We aim to design a low-complexity lossless text compressor using advanced language models. Recent popular LLMs include Transformer~\cite{transformer:2017}, Transformer-XL~\cite{transformerxl:2019}, and RWKV~\cite{rwkv:2023}, and we have conducted extensive experiments on them. 

To explore a low-complexity design, we conduct experiments on various architectures, including Transformer~\cite{transformer:2017}, Transformer-XL~\cite{transformerxl:2019}, and RWKV~\cite{rwkv:2023}. Following the settings in~\cite{LMisComp:2024}, we train models from scratch on enwik8~\cite{hutterprize} and test on enwik9, using a character-based tokenizer and vocabulary size of 128. The prediction capability of language model is largely influenced by context length and too long context increase the running time, so we chunk the data into sequences of $C$ bytes ($C$ is $2048$) similar to~\cite{LMisComp:2024}.

% We train models from scratch using enwik8 and test on enwik9, using a character-based SentencePiece~\cite{spm} tokenizer (SPM-CHAR) with coverage of 0.999 and vocabulary size of 128, which is slightly different from ASCII in~\cite{LMisComp:2024}. decoding typically occurs on devices such as smartphones, so we used the iPhone12 CoreML tool to test the inference time per token. We set the chunk size $C$ to 128 and the batch size to $128$, considering further parallel computing. For models with 169 million parameters, real-time inference is unachievable; therefore, their running times have been omitted. The performance is listed in Table~\ref{cr}. 

\begin{table}[tbp]
    \centering
    \small
    \setlength{\tabcolsep}{5pt} % 调整列间距
    \begin{tabularx}{0.96\columnwidth}{llccc}
    \toprule
    % Chunk & Compressor & \multicolumn{2}{c|}{in-distribution} & \multicolumn{3}{c}{out-of-distribution} \\
    %  & &  5.3x & 7x & 10x & 10.7x & 12x \\
    % & \textasciitilde second
    \textbf{Compressor}  & \textbf{MACs}$\downarrow$ & \textbf{CR(\%)}$\downarrow$ & \textbf{ACR(\%)}$\downarrow$  & \textbf{Time}$\downarrow$ \\
    \midrule
     Tsf-200K &107.78K & 31.86  & 31.86  &8.61ms \\
     Tsf-800K  &412.16K & 27.58 & 27.74  &16.71ms \\
     Tsf-169M &85.11M & 14.81 & 31.71 & -\\
     \cmidrule{1-5}
     TsfXL-200K &155.50K & 27.14  & 27.14  & 4.36ms\\
     TsfXL-800K  &483.25K & 21.49  & 21.65 & 7.57ms\\
     TsfXL-169M  &92.26M & 14.81   & 31.71  & -  \\
     \cmidrule{1-5}
     RWKV-200K  &143.62K & 24.36   & 24.36  & 0.34ms \\
     RWKV-800K  &471.68K & 21.73   & 21.89 & 0.41ms\\
     RWKV-169M  &92.19M & 15.88    &32.78 & - \\   
    \bottomrule
    \end{tabularx}
    \vspace{-2mm}
    \caption{\small Compression performance using Transformer (Tsf), TransformerXL(TsfXL) and RWKV on enwiki9 dataset.}
    \vspace{-5mm}
    \label{tab:comp}
    % \vspace{-2em}
    
\end{table}

To evaluate compression performance, we report both compression ratio (CR) and adjusted compression ratio (ACR). CR is the ratio of the compressed data size to the raw data size, while ACR accounts for model size by adding it to the compressed size, with model size calculated in float16 precision. To assess complexity, we list the Multiply-Accumulate Operations (MACs) and inference time per token on an iPhone12 ANE with a batch size of 128. For models with 169M parameters, real-time inference is unachievable, so their running times are omitted.

Several key findings can be observed from Table~\ref{tab:comp}:
\begin{itemize}
\item \textit{Compression Ratio}:\quad All the models basically achieve lower CRs as the model size increases, with RWKV showing slightly worse scaling than Transformer and TransformerXL. It might be due to the limitation of RWKV's linear attention. Meanwhile, TransformerXL and RWKV are more effective when the model size is relatively small, likely because they both propose explicit memory mechanisms to enlarge the context length.

% In terms of compression performance, all the models basically achieve lower CRs as the model size increases, while the scaling of transformer are slightly better than RWKV. It might result from the linear attention of RWKV, limiting the performance. Meanwhile, TransformerXL and RWKV are more effective when the model size is relatively small, because both propose a explicit memory mechanism to enlarge the context length. 

\item \textit{Complexity}:\quad TransformerXL and RWKV have higher MACs than vanilla Transformers due to their explicit memory designs. However, they exhibit faster inference times thanks to fixed context lengths. Transformers typically process variable sequence lengths, leading to longer inference times as input length increases. While Transformer-XL uses a fixed memory length of 256, RWKV operates with a memory length of 1 and achieves even faster inference speeds.

% In terms of complexity, MACs of TransformerXL and RWKV are higher than vanilla transformer due to the explicit memory design. But, in terms of inference time, they are faster than transformer due to different context length settings. Transformers typically process variable sequence length, resulting the increasing computation complexity along with increasing input length. One chunk with 2048 byte has over 700 tokens and we calculate the total running time and average it. But, the memory length of Transformer-XL is set as a fixed value of 256 and the memory of RWKV is 1 during inference.

\end{itemize}

In summary, RWKV offers the fastest inference speeds with acceptable compression ratios, making it most suitable for our low-complexity design. 

% RWKV models offer an optimal balance of compression performance and model complexity, making them beneficial for practical applications. Therefore, we adopt RWKV as the backbone in our proposed method. 

\subsection{Outlier-aware Tokenizer} \label{sec:3.3}
% TODO: more citations validating pre-compression

% 介绍Tokenizer是一个预压缩的过程
Tokenizers segment text into subword tokens, serving as a pre-compression process \cite{trainingllmsneurallycompressed}. By concealing the character-level composition of each token and merging the most frequent subwords, tokenizers enable the network to model long-distance dependencies and process more data. This approach also enhances the context prediction during inference. Consequently, enhancing tokenizers' performance not only improves the overall compression ratio but also reduces inference complexity.

% Vocab size和Coverage很重要，不用Uniform了
Vocabulary size and coverage value are two important factors for tokenization. Recent language models typically use vocabulary sizes ranging from 32K to 256K tokens. However, a larger vocabulary would increase the inference burden, especially for very small models. While language models usually aim for 100\% character coverage, rare tokens can actually increase the compressed size and reduce the compression efficiency.

\begin{figure}
    \centering
    \includegraphics[width=0.46\textwidth]{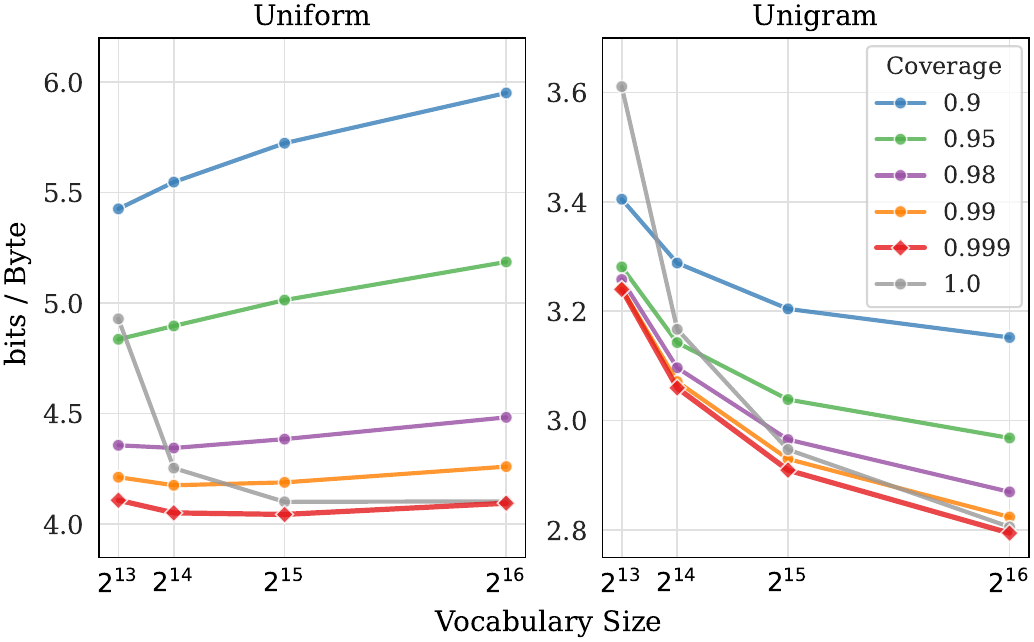}
    % \captionsetup{font=small}
    \vspace{-3mm}
    \captionsetup{font=small}
    \caption{Performance with different coverage values.}
    % \vspace{-3mm}
    \label{fig:bpp}
\end{figure}

To investigate the influence of vocabulary sizes and coverage values, we conduct experiments on enwik8 using Byte Pair Encoding (BPE)~\cite{bpe} as the baseline, which is recognized for its optimal pre-compression performance. We use two distribution methods, ``Unigram" and ``Uniform", to determine the number of bits for each token after tokenization. ``Unigram" calculates the entropy of tokens based on their contextual probabilities in enwik8, assigning shorter codes to more frequent tokens, making it suitable for evaluating in-distribution performance. ``Uniform" assigns identical bits to all tokens, hence is suitable to assess out-of-distribution performance. If the character coverage is lower than 1, there exists a vocabulary set $\Gamma$ containing all known tokens. Text characters excluded by the vocabulary, i.e., the unknown tokens, are identified as outliers and transmitted through a bypass. Therefore, the total compressed size $Bits$ is defined as
\begin{equation}
    Bits = H_{x_i \in \Gamma}(p) +  R (x_i \notin \Gamma)
    % TODO[Done]: 左右都是 R, propose 左/右 换个符号
\end{equation}
where $H(p)$ represents the compressed size of known tokens, as defined in Eq.(\ref{eq.1}), and $R(x_i \notin \Gamma)$ denote the consumed bits for outliers to be coded directly as a UTF-8 byte sequence. Two parts contribute to their total bit sizes. The ``bits per byte (bpb)" metric (the number of bits to encode one byte of text) is then calculated to assess the tokenizer's pre-compression efficiency, with a lower value indicateing better performance, as shown in \cref{fig:bpp}.

% where $H(p)$ is defined in Eq.(\ref{eq.1}). Then by calculating $R$ divided by the raw byte size, we can evaluate "bits per byte (bpb)" (the number of bits to encode one byte of the text) in \cref{fig:bpp}, where the lowest value indicates the best. 

% "Unigram" calculates the entropy of tokens based on their contextual probabilities in enwik8, assigning shorter codes to more frequent tokens to evaluate in-distribution performance. "Uniform" assigns identical bits to all tokens to evaluate out-of-distribution performance.

% TODO[Done]: 这个表放在左边没顶满的文字上面是不是有点丑
\begin{table}[tb]
    \centering
    \small
     % \renewcommand{\arraystretch}{1.2}
     % \captionsetup{font=small}
     \vspace{-2mm}
    \setlength{\tabcolsep}{6pt} % 调整列间距
    \begin{tabularx}{0.96\columnwidth}{lcccc}
    \toprule
     Vocabulary Size  & 8K & 16K & 32K  & 64K \\
     \midrule
     NPU (bs=16) & 0.36ms & 0.51ms & 2.44ms  & 3.96ms\\
     CPU (bs=1)  & 0.12ms & 0.23ms & 0.52ms  & 1.01ms\\
    \bottomrule
    \end{tabularx}
    % \captionsetup{font=small}
    \vspace{-2mm}
    \captionsetup{font=small}
    \caption{Decoding speeds using different vocabulary sizes.}
    \label{tab:dtime}  
    \vspace{-4mm}
\end{table}

It can be observed the coverage of $0.999$ achieves the lowest bpb for both ``Uniform" and ``Unigram" methods. When using the ``Unigram" method, increasing the vocabulary size can reduce the sequence length, packing more information into the context and resulting in a better compression ratio. However, this benefit diminishes if the evaluation data's distribution differs from the training data, as measured using the ``Uniform" method. In other words, the largest vocabulary size is not always optimal. Moreover, our evaluation using RWKV-800K (Table~\ref{tab:dtime}) shows that the inference time increases with the vocabulary size. Thus, a relatively small vocabulary size is essential for a low-complexity design.

\begin{figure}[ht]
\centering
\includegraphics[width=1.0\columnwidth]{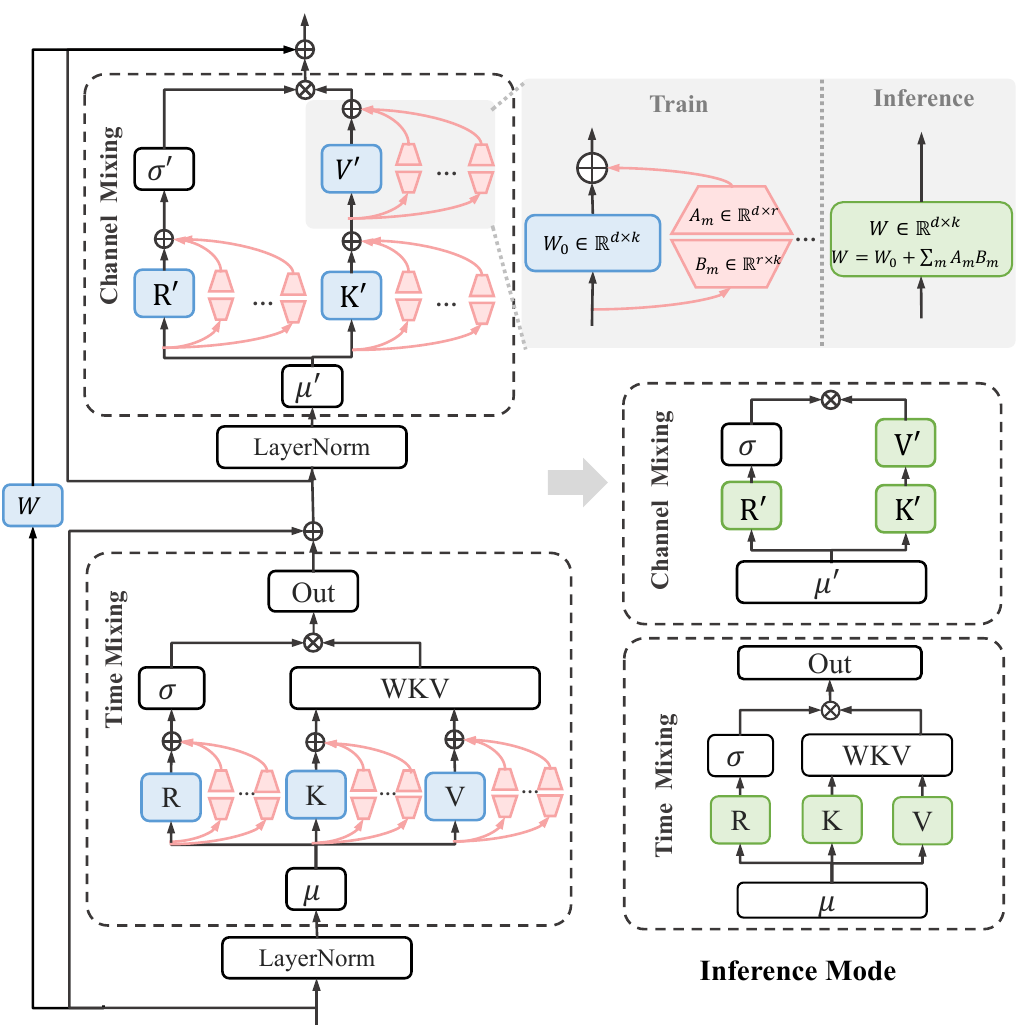} %
% \captionsetup{font=small}
\vspace{-6mm}
\captionsetup{font=small}
\caption{Proposed high-rank reparameterization method.}
\vspace{-3mm}
\label{fig:rwkv}
\end{figure}

\subsection{High-rank Reparameterization} \label{sec:3.4}

% TODO[Done]: 这个 HIRA 的缩写是在这里提呢还是在比较靠前的地方呢? 暂时先放这里好了 
To enhance the inference performance of RWKV without compromising the computational efficiency, we introduce a High-rank (HiRA) reparameterization strategy. As shown in \cref{fig:rwkv}, we add additional branches for each R, K, V layer in both ``Channel Mixing" and ``Time Mixing" modules during training to enhance the training effectiveness, and then merge them into the main branches during inference to reduce the number of parameters. Inspired by Low-rank adaption (LoRA)~\cite{lora:2021}, we increase the parameters of these additional branches through matrix decomposition, thereby improving their capability. 

Specifically, assuming the original parameters of the R, K, V layers are $W_0\in \mathbb{R}^{d\times k}$, we do not directly add several parallel branches, such as 1x1 convolution or shortcuts as in \cite{ding2021repvgg}. Instead, given the $m$-th branch, we decompose it into the product of two high-rank matrices: $A_{m}\in \mathbb{R}^{d\times r}$ and $B_{m}\in \mathbb{R}^{r\times k}$. Therefore, during training, these layers' output is obtained by summing the outputs of the main branch and several bypass branches. Unlike LoRA, the matrices $W_0$, $A_m$, and $B_m$ are optimized simultaneously. The effect of the rank $r$ and the number of branches $m$ will be discussed in the ablation studies. 

As illustrated in the right part of \cref{fig:rwkv}, during inference, the additional branches $A_m\times B_m$ are merged into the main branch via structural reparameterization:
\begin{equation}
    W = W_0 + \sum_{m} A_m\times B_m
\end{equation}
Such a single-path structure after reparameterization ensures lower running time and memory usage during inference. Despite the structural changes, the multi-branch parameters are retained, ensuring high inference performance. Besides, we add a linear layer (depict as $W$) across the whole RWKV module to further enhance the performance with negligible computational overhead.

\section{Experiments}

\subsection{Experimental Setup}

% For RWKV models, we adjust the number of layers, {\color{blue} attention sizes in the Time Mixing module, and hidden sizes in the Channel Mixing module} 

% TODO[Done]: attn size / hidden 是否就是 attention embedding dimension -> 好像 embed dim 更容易理解因为 RWKV 用的是这种说法. -> 但都用 embed dim 怎么区分 time mixing 和 channel mixing 没想好.
\paragraph{Implementation Details.} For RWKV models, we adjust the number of layers, attention embedding dimension and hidden sizes to achieve target model sizes. We train the models using the AdamW~\cite{adamw} optimizer with an initial learning rate of 1e-4 and a linear learning rate scheduler with a decay rate of 0.999 over 20 epochs, without warm-up. All the models are trained with a sequence length of 2048 bytes and a batch size of 64.

% \begin{table}[h]
%     \centering
%     \small
%      % \renewcommand{\arraystretch}{1.2}
%     % \setlength{\tabcolsep}{7pt} % 调整列间距
%     \captionsetup{font=small}
%     \caption{Model structures across different sizes.}
%     \vspace{-3mm}
%     \begin{tabularx}{0.95\columnwidth}{cccc}
%     \toprule
%     \textbf{Model} & \textbf{Layer} & \textbf{Embedding dim} & \textbf{Hidden size} \\
%     \midrule
%     L3TC-200K & 2 & 96 & 96 \\
%     L3TC-800K & 2 & 176 & 192 \\
%     L3TC-3.2M & 3 & 256 & 512 \\
%     L3TC-12M  & 4 & 384  & 1024 \\
%     \bottomrule
%     \end{tabularx}
%     \label{net}
%     % \vspace{-2em}
% \end{table}

During evaluation, for learned compressors, we split the data into sequences of $2048$ bytes and enable parallel batch processing. For traditional compressors with context lengths exceeding 2048 bytes, chunking can deteriorate the compression ratio. Therefore, we use unchunked data to maintain their optimal performance.

% The configurations of our model include the use of AdamW optimizer with an initial learning rate of 1e-4 and a linear learning rate scheduler with decay rate of 0.999 every 10 steps. We train all of our model using content length of 2048 and batch size of 64. The training process consists of 20 epoch without warm-up.
% \vspace{-1mm}
\paragraph{Dataset and Metrics.} The \emph{enwik8/enwik9} datasets, consisting of the first 1GB and 100MB of the English Wikipedia XML, are commonly used to evaluate text compression performance. Since enwik8 only contains 10\% data of enwik9, they represent a significant distribution shift. Therefore, we train our L3TC models on enwik8 and evaluate them on both enwik8 and enwik9 to assess the in-distribution and out-of-distribution compression performance. 

% Since enwik8 contains the first 100MB of enwik9. 
% We also include \emph{text8} dataset, a filtered subset of enwik9, to check the performance on clean text.

% The compression performance is evaluated using CR and ACR metrics, similar to \cref{sec:3.2}. We use MACs and decoding speeds to report the complexity. 

We evaluate compression performance using CR and ACR metrics, similar to \cref{sec:3.2}, and assess complexity through MACs and decoding speeds. The decoding speeds are measured on typical computing platforms, including server GPUs (NVIDIA A100 80 GB) and device NPUs (iPhone12 Apple Neural Engine). To measure on-device performance, we convert models to CoreML packages~\cite{coreml} and use Xcode software to get the inference time. 

% By simultaneously utilizing the CPU, GPU, and Apple Neural Engine (ANE, i.e. NPU), Core ML enhances the inference speed while minimizing memory usage and power consumption. The efficiency gains from Core ML are more pronounced with larger model sizes or batch sizes. 

% CR is defined as the ratio of compressed data size to the raw data size, while ACR takes the compressor's size into consideration and adds it to the compressed data size. Besides, we evaluate each compressor's complexity using Multiply-Accumulate Operations (MACs) and running time. 

% \vspace{-1mm}
\paragraph{Baselines.} Comparisons are conducted with classical compressors including gzip~\cite{gzip}, bzip2~\cite{bzip2}, zstd~\cite{zstd}, and learned compressors based on pretrained language models, such as Llama2~\cite{llama2}, Chinchilla~\cite{chinchilla}, and RWKV~\cite{rwkv:2023}. Models with various sizes are downloaded from Huggingface. We omit LLMZip~\cite{llmzip:2023} as its results are equivalent to Llama-7B. Models in tszip~\cite{tszip} are also included in comparisons. Since Chinchilla models are not open-source, their experimental results are derived from the paper \cite{LMisComp:2024}. All other models' results are recorded based on our implementations. We also compare the results of transformer in~\cite{LMisComp:2024}, which are slightly different from~\cref{tab:comp}, because the network structures and tokenizers are not exactly the same.

% Our comparison is primarily with the learned compression methods. Specifically, we evaluate not only the vanilla decoder-only transformers trained on enwik8, but also the pretrained large language models like Llama2~\cite{llama2}, Chinchilla~\cite{chinchilla}, and RWKV~\cite{rwkv:2023}. Models of different sizes are compared to assess the relationship between compression capability and model size. 

% LLMZip~\cite{llmzip:2023} and TSZip~\cite{tszip} are also included in comparisons. Since the Chinchilla models are not open-sourced, their experimental results are derived from the paper \cite{LMisComp:2024}. All other models' results are recorded based on our implementations. We also compare the performance of general-purpose traditional lossless compressors: GZIP, BZIP2, and ZSTD. However, advanced compressors such as CMIX~\cite{cmix} and NNCP~\cite{nncp} take approximately 7.2 and 2.8 days to decode 1GB of data, respectively. Such an excessive decoding complexity is unacceptable for practical usage. Therefore, we exclude them from the comparison.

% advanced traditional compressors such as CMIX and NNCP exhibit decoding speeds of 10 min/MB and 4 min/MB, respectively, which means that they require approximately 7.2 and 2.8 days to decompress the data of 1GB.

% Compression Ratios (CR) Using different models on enwiki9 dataset with 1GB text, where bold text indicates the best performance, while underlined text indicate the second best.
\begin{table*}[tbp]
    \centering
    \small
     \renewcommand{\arraystretch}{1.2}
    \setlength{\tabcolsep}{7pt} % 调整列间距
    \vspace{-3mm}
    \begin{tabularx}{0.96\textwidth}{llllll}
    \toprule
    % Chunk & Compressor & \multicolumn{2}{c|}{in-distribution} & \multicolumn{3}{c}{out-of-distribution} \\
    %  & &  5.3x & 7x & 10x & 10.7x & 12x \\
    % & \textasciitilde second
    \textbf{Chunk} & \textbf{Compressor} & \textbf{Tokenizer} & \textbf{MACs} & \textbf{CR(\%)} & \textbf{ACR(\%)}   \\
    \midrule
     default & gzip &  - & - & 32.26  &  32.26 \\ %& \textasciitilde seconds\\
      & bzip2 &  - & - & 25.40 &  25.40 \\ %& \textasciitilde seconds  \\
     & zstd (-22 --ultra)  &  - & - & 21.49 &  21.62 \\ % & \textasciitilde seconds \\
     & $\text{tszip}^{\textbf{*}}$~\cite{tszip}  & RWKV-Pile-Tokenizer-50K & 130.71M  & 13.54 & 47.34 \\ % & \\
     % & CMIX v20 (lstm) &  - & - & 10.99 &  \textbf{10.99} & 7.2d \\
     % & NNCP v3.2 (transformer) &  - & - & 10.66 &  \textbf{10.66} & 2.8d \\
     \cmidrule{1-6}
     % 2048 & Transformer-200K &SPM-Char-128 &107.78K & 31.86  & 31.86  \\
     % & Transformer-800K &SPM-Char-128 &412.16K & 27.58 & 27.74 \\
     % & Transformer-169M &SPM-Char-128 &85.11K & 14.81 & 31.71\\
     % \cmidrule{2-7}
     % & TransformerXL-200K &SPM-Char-128 &155.50K & 27.14  & 27.14  & 4.36ms/bs=128\\
     % & TransformerXL-800K &SPM-Char-128 &483.25K & 21.49  & \underline{21.65} & 7.57ms/bs=128\\
     % & TransformerXL-169M &SPM-Char-128 &92.26M & 14.81   & 31.71  &  \\
     % \cmidrule{2-7}
     % & RWKV-200K &SPM-Char-128 &143.62K & 24.36   & 24.36  & 0.34ms/bs=128 \\
     % & RWKV-800K &SPM-Char-128 &471.68K & 21.73   & \underline{21.89} & 0.41ms/bs=128\\
     % & RWKV-169M &SPM-Char-128 &92.19M & 15.88    &32.78 & \\  
     % \cmidrule{2-7}
     2048 & $\text{Llama2-7B}^{\textbf{*}}$  & Llama-tiktoken-32K & 6.61G  & 8.62   &1408.62  \\ % &\textasciitilde days\\
     & $\text{Llama2-13B}^{\textbf{*}}$  & Llama-tiktoken-32K & 12.85G & \textbf{8.04} {\small({-75\%})}  &2608.04 \\ %&\textasciitilde days\\
     & $\text{Chinchilla-1B}^{\textbf{*}}$~\cite{LMisComp:2024}  & Chinchilla-tokenizer & - & 11.30   &211.30  \\ %\textasciitilde days\\  
     & $\text{Chinchilla-7B}^{\textbf{*}}$~\cite{LMisComp:2024}  & Chinchilla-tokenizer & - & 10.20   &1410.20 \\ %\textasciitilde days\\  
     & $\text{Chinchilla-70B}^{\textbf{*}}$~\cite{LMisComp:2024}  & Chinchilla-tokenizer & -  & 8.30  &14008.30 \\ %& \textasciitilde days\\
     % & $\text{RWKV-169M}^{\textbf{*}}$ & RWKV-Pile-tokenizer-50K & 130.71M & 14.23 & 48.03 \\ %&  \\
     % & $\text{RWKV-430M}^{\textbf{*}}$ & RWKV-Pile-tokenizer-50K & 378.84M & 12.38 & 98.38 \\ %&  \\
     & $\text{RWKV-1.5B}^{\textbf{*}}$ & RWKV-Pile-tokenizer-50K & 1.41G   & 10.89 & 310.89 \\ %&  \\
     % & $\text{RWKV-3B}^{\textbf{*}}$ & RWKV-Pile-tokenizer-50K & 2.86G & 10.33 & 610.33   \\ %\\
     & $\text{RWKV-7B}^{\textbf{*}}$ & RWKV-Pile-tokenizer-50K & 7.19G & 9.68 &  1409.68  \\ %\\
     % & $\text{LLMZip}^{\textbf{*}}(\footnotesize{\text{zlib}})$~\cite{llmzip:2023}  &Llama-tiktoken-32K  &6.61G  &13.11 &1413.11 \\ %& \\
     % \cmidrule{2-6}
     \cmidrule{2-6}
     & Tsf-200K~\cite{LMisComp:2024} & ACSII  & -  & 30.90 & {30.90}  \\
     & Tsf-800K~\cite{LMisComp:2024} & ACSII  & -  & 21.70 & {21.86} \\
     & Tsf-3.2M~\cite{LMisComp:2024} & ACSII  & -  & 17.00 & 17.64  \\
     % & Tsf-200K & SPM-Char-128-0.999  & 107.78K  & 31.86 {\small({-1\%})} & {31.86} {\small({-1\%})} \\ %& \\
     % & Tsf-800K & SPM-Char-128-0.999  & 412.16K  & 27.58 {\small({-15\%})} & {27.74} {\small({-14\%})}\\ % & \\
     % & Tsf-169M & SPM-Char-128-0.999  & 85.11M  & 14.81 {\small({-54\%})} & \underline{31.71} {\small({-2\%})} \\ %& \\
     & \emph{\textbf{L3TC-200K}} & Outlier-aware tokenizer  & \textbf{1.72M} & 18.48 {\small({-43\%})} & {18.48} {\small({-43\%})} \\ %& \\
     & \emph{\textbf{L3TC-800K}} & Outlier-aware tokenizer  & 3.33M & 17.56 {\small({-46\%})} & {17.72} {\small({-45\%})} \\ %& \\
     & \emph{\textbf{L3TC-3.2M}} & Outlier-aware tokenizer & 6.17M &16.23 {\small({-50\%})} & \textbf{16.87} {\small({\color{blue}\textbf{-48\%}})} \\ %& \\
     & \emph{\textbf{L3TC-12M}} & Outlier-aware tokenizer  & 12.99M &16.00 {\small({-50\%})} & 18.40 {\small({-43\%})} \\ %& \\
    \bottomrule
    \end{tabularx}
    \vspace{-2mm}
    \captionsetup{font=small}
    \caption{Compression ratios on enwiki9 dataset (1GB): Bold indicates best performance. Although pretrained Llama2-13B achieves the best performance with the lowest CR, it is impractical to use. When considering the overhead of model size, our proposed \emph{L3TC-3.2M} achieves the best performance with the lowest ACR, saving 48\% bits when using \emph{gzip} as an anchor.}
    \label{cr2}

\end{table*}

\begin{table}[tbh]
    \centering
    \small
    \setlength{\tabcolsep}{5pt} % 调整列间距
    \vspace{-2mm}
    \begin{tabularx}{0.98\columnwidth}{llll}
    \toprule
    \textbf{Chunk} & \textbf{Compressor} & \textbf{CR(\%)} & \textbf{ACR(\%)}   \\
    \midrule
    default & gzip  & 36.45  &  36.45 \\
    & tszip~\cite{tszip}   & 13.83 & 351.83  \\
    \cmidrule{1-4}
    % 2048 & $\text{LLMZip}^{\textbf{*}}$~\cite{llmzip:2023} \\ 
    2048 & L3TC-200K & 17.55 {\small({-52\%})} &  17.95 {\small({-51\%})}\\
    & L3TC-800K & 15.68 {\small({-57\%})} &  \textbf{17.28} {\small({-53\%})}\\
    & L3TC-3.2M & 13.81 {\small({-62\%})} &  20.23 {\small({-44\%})} \\
    & L3TC-12M  & \textbf{11.24} {\small({-69\%})}  & 35.24 {\small({-3\%})}\\
    % LLMZip  & - & - \\
    \bottomrule
    \end{tabularx}
    \vspace{-2mm}
    \captionsetup{font=small}
    \caption{Compression ratios on enwik8 dataset (100MB).}
    \label{cr3}
    \vspace{-4mm}
\end{table}

\subsection{Compression Performance}
% \subsection{Comparison Results}
Table~\ref{cr2} illustrates the compression ratios for different compressors on enwik9. Models labeled with $\textbf{*}$ are pretrained models on larger text datasets using their own tokenizers. Models without $\textbf{*}$ are trained from scratch on enwiki8. Our proposed L3TC models are highlighted in bold and in italics. 

As shown in Table~\ref{cr2}, classical compressors, e.g., gzip, offer moderate compression ratios with minimal computational complexity. Compressors based on pretrained models (Llama2, Chinchilla, and RWKV) achieve superior compression performance, with Llama2-13B leading by achieving a 75\% bit saving compared to gzip. However, these models suffer from significant model sizes (ranging from 169M to 70B parameters) and excessive resource requirements. In fact, using such large models to compress small-scale data is highly inefficient, with ACR values ranging from 48\% to 14,000\%. Even compressed using low-bit quantization, the model size overhead remains substantial, making them impractical for real-world deployments. 

By contrast, the proposed \emph{L3TC} models have smaller model sizes and lower computational complexity. \emph{L3TC-3.2M} achieves 50\% bit saving compared to gzip and outperforms other learned compressors by saving 48\% more bits when accounting for model size. \emph{L3TC} also offers compression ratios comparable to other learned compressors, with a $50\times$ reduction in model parameters, as shown in \cref{fig1:teaser}.

Table~\ref{cr3} records the compression ratios on enwik8, providing an assessment of in-distribution performance. While larger model sizes generally enhance compression, L3TC-12M shows no further gains on enwik9 compared to L3TC-3.2M. Actually, when the model size approaches the size of training data, the model is likely to overfit to the training set. Therefore, for general usage, L3TC-12M is excluded in the following section.

% It can be observed the compression performance basically improves with the increase of model size, but L3TC-12M does not bring further improvement on enwik9 compared to L3TC-3.2M. Indeed, when the model size is comparable to the size of the training data, the model is likely to overfit to the training set. Therefore, for general usage, L3TC-12M should be excluded. 

% {\color{red} Add results on text8 dataset}

% Hence these models' ACR values are quite similar to the CR values. We explore their compression performance under three model sizes: 200K, 800K, and 3.2M. 

% Transformer models with smaller sizes exhibit compression ratios slightly better than traditional compressors, with improvements as the model size increases. However, our proposed L3TC models are 

% TODO: 这个a,b,c小标题要不也去掉算了 -> 感觉去掉也看得懂 => 或者小标题改一改 - 待定.

% 改成了和图里对应的 decoding speed
\subsection{Decoding Speed}
% \subsection{Comparison of Running Time}

%TODO: Add the running time Table (unit: MB/s, KB/s etc.)

% We test the decoding speed not only on mobile devices but also on servers (Linux). 
% TODO: linux 设备配置 -> cpu, gpu; Linux 解码时间
% To measure on-device performance, we convert the models to Core ML packages~\cite{coreml} and test them on an iPhone 12. By simultaneously utilizing the CPU, GPU, and Apple Neural Engine (ANE, i.e. NPU), Core ML enhances the inference speed while minimizing memory usage and power consumption. The efficiency gains from Core ML are more pronounced with larger model sizes or batch sizes. 

% The unmodified RWKV models are adopted for comparison. 

Table~\ref{decodetime} compares the decoding speeds for various methods. To calculate the decoding speed (bytes per second), we multiply the batch size by the average byte length per token and then divide the result by the batch inference time. The average byte lengths for the character-based tokenizer, Llama tiktokenizer, RWKV-pile tokenizer, and our proposed outlier-aware tokenizer are $1$, $2.94$, $3.37$ and $3.29$, respectively. On the A100, the batch size is 2048 by default or the maximum size without out-of-memory. When running on mobile devices, the batch size is set to 256. Pretrained models with hundreds of millions to billions of parameters are difficult to run on devices with limited memory and computational resources, so their decoding speeds are not reported.

\begin{table}[tb]
    \centering
    \small
     \renewcommand{\arraystretch}{1.0}
    \setlength{\tabcolsep}{7pt} % 调整列间距
    \vspace{-2mm}
    \begin{tabularx}{0.98\columnwidth}{lrr}
    \toprule
    \textbf{Compressor} & \textbf{iPhone12@ANE} & \textbf{GPU@A100}   \\
    \midrule
    % TSZip  & - &  \\ 
    $\text{Llama2-7B}^{\textbf{*}}$(\small{LLMZip}) & - & 280 B/s \\
    $\text{Llama2-13B}^{\textbf{*}}$ & - & 85 B/s \\
    % $\text{RWKV-430M}^{\textbf{*}}$  & - & 66 KB/s\\
    $\text{RWKV-1.5B}^{\textbf{*}}$  & - & 20 KB/s\\
    % $\text{RWKV-3B}^{\textbf{*}}$  & - & 10 KB/s\\
    $\text{RWKV-7B}^{\textbf{*}}$  & - & 4 KB/s\\
    % LLMZip  & - & \\
    tszip~\cite{tszip}  & - & 180 KB/s \\
    \midrule
    Tsf-200K   & 15 KB/s  & 48 KB/s  \\
    TsfXL-200K & 29 KB/s  & 170 KB/s \\
    RWKV-200K & 750 KB/s &  1.32 MB/s\\
    % RWKV-800K & 1.3 MB/s  &  \\
    \textbf{L3TC-200K} & \textbf{1.30 MB/s} & \textbf{4.35 MB/s} \\
    \textbf{L3TC-800K} & 980 KB/s &  3.87 MB/s\\
    \textbf{L3TC-3.2M} & 633 KB/s &  2.50 MB/s\\
    \bottomrule
    \end{tabularx}
    \vspace{-2mm}
    \captionsetup{font=small}
    \caption{Decoding speed on different platforms.}
    \label{decodetime}
    \vspace{-4mm}
\end{table}

% Decoding speeds (byte per second) are calculated by multiplying the batch size by the average byte length per token, divided by the batch-inference time.

% After tokenization, one token consumes 3.3 bytes in average. 

Table~\ref{decodetime} shows that the Llama models operate at speeds measured in bytes per second (B/s), while the pretrained RWKV models are significantly faster, measured in kilobytes per second (KB/s). Our proposed methods \emph{L3TC} are the fastest, reaching decoding speeds up to 1.30 MB/s on mobile devices. The speed gains of \emph{L3TC}-200K over the default RWKV-200K are primarily due to our proposed outlier-aware tokenizer. This tokenizer merges frequent subwords, allowing more data to be processed in a single inference, whereas the default RWKV-200K uses character-based tokenizer and processes only one character at a time.

% {\color{blue} The speed gains of \emph{L3TC}-200K over default RWKV-200K are primarily due to the proposed outlier-aware tokenizer, whose average byte length per token is longer than character-based tokenization.}
% TODO[Done]: 这句话有点没太懂, "average byte length per token" 是每个 token 的 bits 数吗 -> 和速度的关系是? -> propose 小修一下这句话.

\begin{figure}[tb]
    \centering
    \captionsetup[subfigure]{labelformat=empty}
    \subfloat[]{\includegraphics[height=1.8in]{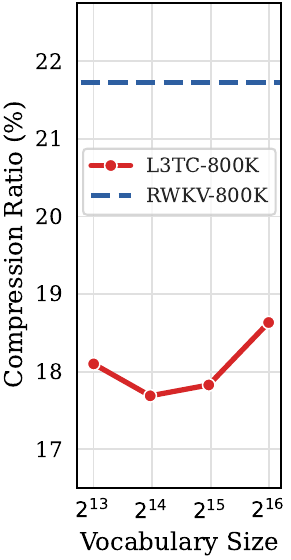}%
    \label{vocab}}
    \hfill
    \subfloat[]{\includegraphics[height=1.8in]{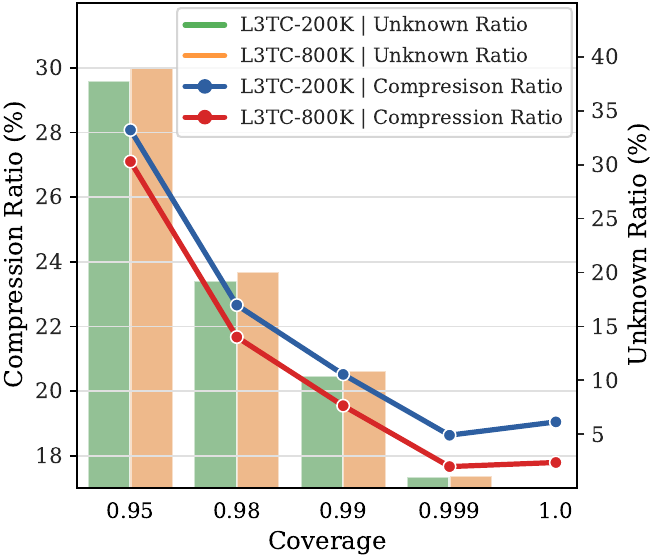}%
    \label{coverage}}
    % \captionsetup{font=small}
    \vspace{-6mm}
    \captionsetup{font=small}
    \caption{Discussion on the outlier-aware tokenizer.}
    \vspace{-3mm}
    \label{fig:abla_tokenizer}
\end{figure}

\subsection{Ablation Study}

\paragraph{Discussion on the Outlier-aware Tokenizer.}

We conduct experiments with different vocabulary sizes and coverage settings. As shown in the left sub-figure of \cref{fig:abla_tokenizer} (the coverage for SPM-BPE is consistently set at 0.999), the compression ratios on enwik9 decrease as vocabulary size increases, adhering to the scaling law~\cite{scalinglaw}. However, excessively large vocabularies may require a larger model capacity, leading to a slight decline in compression performance when vocabulary sizes exceed 16K. Therefore, we empirically set the vocabulary size to 16K.

The right sub-figure of \cref{fig:abla_tokenizer} demonstrates the relationship between coverage values and compression ratios using line graphs. The bar charts, whose y-axis is depicted on the right, indicate the proportion of unknown tokens' size to the compressed data size (i.e., $\frac{8 \times N (x_i \notin \Gamma}{R}$). As the coverage value increases, both unknown ratios and compression ratios exhibit a decreasing trend. However, the compression ratio reaches the lowest point when the coverage reaches 0.999, after which it slightly increases. This results validate the effectiveness of proposed outlier-aware tokenizer.

\vspace{-1mm}
\paragraph{Discussion on High-rank Reparameterization.}

% \subsubsection{4.4.2\quad Discussion on High-Rank Reparameterization}

% TODO: Suggest to put Fig.6 （b）(c)(d) 一张图，讨论 High-Rank Multi-Branch Reparam的影响；

% TODO: Fig.6(a) 想想再加2张图凑成3个，讨论Tokenizer的影响，分别是unknown token (outlier) size的占比（补一张），以及 RWKV-800K-不同vocab的压缩率(现在的图），耗时（补一张）; 

% We conduct experiments on the effect of the vocabulary size based on RWKV-800K backbone, and the results are listed in Fig~\ref{vocab}. We utilize the sentence piece~\cite{spm} with the vocabulary size of 128, $1K$, $2K$, $4K$, $16K$, $32K$, $64K$ to explore the influence on the compression ratio.  Here, we use raw compression ratio instead of adjusted CR as metrics since the model size is fixed and $800K$ parameters are relatively small compared to compressed size. 

\begin{figure}[tb]
    \centering
    % \subfloat[Discussion on ranks]{\includegraphics[height=1.75in]{Figure/rank_cr.pdf}
    % \label{rank}}
    % \hfill
    % \subfloat[Discussion on branches]{\includegraphics[height=1.75in]{Figure/branch_cr.pdf}
    % \label{branch}}
    % \hfill
    \subfloat[\small Effect of HiRA's settings.]{\includegraphics[height=1.75in]{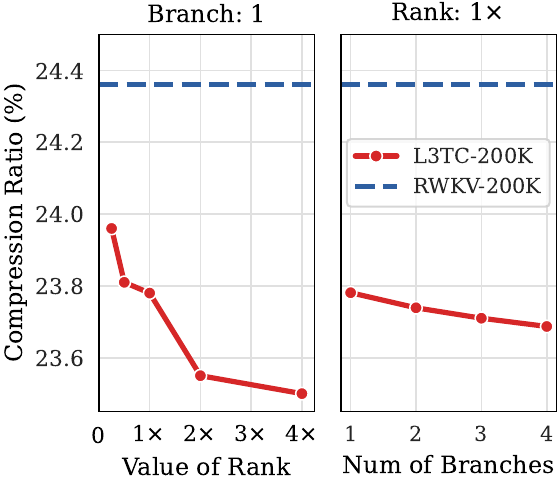}
    \label{rank}}
    \hfill
    \subfloat[\small Effect of HiRA.]{\includegraphics[height=1.75in]{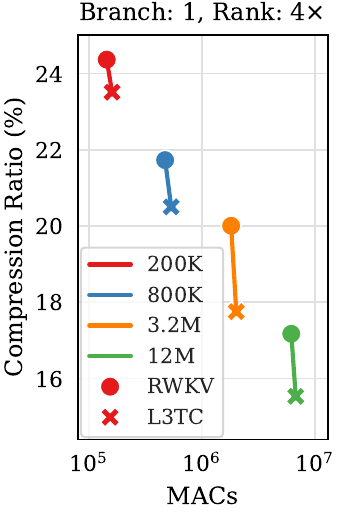}
    \label{modelsize}}
    % \captionsetup{font=small}
    \vspace{-2mm}
    \captionsetup{font=small}
    \caption{Discussion on high-rank reparameterization.}
    \label{fig:abla_hira}
    % \vspace{-3mm}
\end{figure}

To examine the effect of rank $r$ and the number of branches $m$, we conduct experiments as \cref{rank}. The left subfigure illustrates that given $r$ ranging in the set of \{$0.25$, $0.5$, $1$, $2$, $4$\}, compression ratio continuously decrease along with the increase of ranks. We use $r$ as 4 in our experiments. The right subfigure shows the compression ratios decrease minimally with the increase of more additional branches. Therefore, we use only one branch in our experiments.

To further validate the effectiveness of the proposed high-rank reparameterization, we conduct experiments with default character-based tokenizers in \cref{modelsize}. The circles refer to the original unmodified RWKV models, while the crossed points represent the proposed L3TC models using HiRA with one branch and the rank of $4$. Points of the same color correspond to models with equivalent sizes. Obviously, introducing HiRA significantly improves compression performance. It is noteworthy that reparameterization itself does not increase MACs; the marginal increase is due to the added linear shortcut parallel to the whole RWKV module.

\begin{figure}[tb]
\centering
\includegraphics[width=0.9\columnwidth]{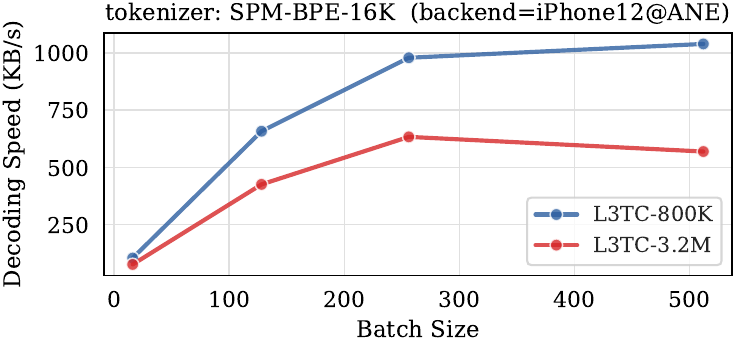} %
% \captionsetup{font=small}
\vspace{-2mm}
\captionsetup{font=small}
\caption{Decoding speeds with different batch sizes.}
\vspace{-4mm}
\label{fig:batchsize}
\end{figure}

\vspace{-1mm}
\paragraph{Discussion on Decoding Speed.}

Batch size is a key factor to decoding speed but large batch size can not always improve the decoding speed proportionally, especially on resource-limited devices. We analysis the effect of batch size on iPhone12. As shown in Fig.~\ref{fig:batchsize}, when batch size is relatively small, the decoding speed is proportion to the batch size, but when the computational resources saturate (i.e. batch size is larger than 256), the decoding speed is not increased any more. Thus, we use a batch size of 256 in our settings when running on device.

% \noindent\textbf{Effectiveness of HiRA re-parameterization}  We validate the effectiveness of our proposed HiRA re-parameterization with different model sizes and complexity in Fig~\ref{modelsize}.

% \begin{figure}[htb]
% \centering
% \includegraphics[width=0.9\columnwidth]{tc_token_1.pdf} % Reduce the figure size so that it is slightly narrower than the column. Don't use precise values for figure width.This setup will avoid overfull boxes.
% \caption{Compression Ratio with different tokens.}
% \label{fig1}
% \end{figure}

% \vspace{-3mm}
\section{Conclusion}

In this paper, we propose a novel learned lossless low-complexity text compressor (L3TC) method. First, we conduct extensive experiments to compare various architectures and select RWKV as our backbone due to its fast decoding speed. Second, we propose an outlier-aware tokenizer that focuses on frequent tokens while bypassing infrequent outliers. Third, we propose a high-rank reparameterization strategy to enlarge the model capacity during training while not increasing inference complexity. Experiments demonstrate that our proposed L3TC achieves impressive lossless compression performance, with 48\% size reductions compared to gzip. \emph{L3TC} offers compression performance comparable to other learned compressors with $50\times$ model size reductions. Besides, L3TC is the fastest among all the learned compressors, with a decoding speed of up to MB/s.

\section{Acknowledgments}

This work was supported by the Ant Group Research Fund, the Fundamental Research Funds for the Central Universities, the National Natural Science Foundation of China (62102024, 62331014, 62431015), and the Shanghai Key Laboratory of Digital Media Processing and Transmissions, China.

% \section{Preparing an Anonymous Submission}

% This document details the formatting requirements for anonymous submissions. The requirements are the same as for camera ready papers but with a few notable differences:

% \begin{itemize}
%     \item Anonymous submissions must not include the author names and affiliations. Write ``Anonymous Submission'' as the ``sole author'' and leave the affiliations empty.
%     \item The PDF document's metadata should be cleared with a metadata-cleaning tool before submitting it. This is to prevent leaked information from revealing your identity.
%     \item References must be anonymized whenever the reader can infer that they are to the authors' previous work.
%     \item AAAI's copyright notice should not be included as a footer in the first page.
%     \item Only the PDF version is required at this stage. No source versions will be requested, nor any copyright transfer form.
% \end{itemize}

% You can remove the copyright notice and ensure that your names aren't shown by including \texttt{submission} option when loading the \texttt{aaai25} package:

% \begin{quote}\begin{scriptsize}\begin{verbatim}
% \documentclass[letterpaper]{article}
% \usepackage[submission]{aaai25}
% \end{verbatim}\end{scriptsize}\end{quote}

\bibliography{aaai25}

\begin{thebibliography}{40}
\providecommand{\natexlab}[1]{#1}

\bibitem[{Apple(2023)}]{coreml}
Apple. 2023.
\newblock {Guide to Core ML tools}.
\newblock \url{https://apple.github.io/coremltools/docs-guides/index.html#}.

\bibitem[{Bellard(2021)}]{nncpv2:2021}
Bellard, F. 2021.
\newblock {NNCP v2: Lossless data compression with transformer.}
\newblock \emph{Technical report, Amarisoft}.

\bibitem[{Bellard(2023)}]{tszip}
Bellard, F. 2023.
\newblock {ts\_zip: Text Compression using Large Language Models}.
\newblock \url{https://bellard.org/ts_zip/}.

\bibitem[{Boutell(1997)}]{png}
Boutell, T. 1997.
\newblock \emph{PNG (Portable Network Graphics) Specification Version 1.0}.
\newblock W3C.

\bibitem[{Dai et~al.(2019)Dai, Yang, Yang, Carbonell, Le, and Salakhutdinov}]{transformerxl:2019}
Dai, Z.; Yang, Z.; Yang, Y.; Carbonell, J.; Le, Q.~V.; and Salakhutdinov, R. 2019.
\newblock Transformer-XL: Attentive Language Models Beyond a Fixed-Length Context.
\newblock arXiv:1901.02860.

\bibitem[{Delétang et~al.(2024)Delétang, Ruoss, Duquenne, Catt, Genewein, Mattern, Grau-Moya, Wenliang, Aitchison, Orseau, Hutter, and Veness}]{LMisComp:2024}
Delétang, G.; Ruoss, A.; Duquenne, P.-A.; Catt, E.; Genewein, T.; Mattern, C.; Grau-Moya, J.; Wenliang, L.~K.; Aitchison, M.; Orseau, L.; Hutter, M.; and Veness, J. 2024.
\newblock {Language Modeling Is Compression}.
\newblock In \emph{{Proceedings of International Conference on Learning Representations (ICLR)}}.

\bibitem[{Deng et~al.(2009)Deng, Dong, Socher, Li, Li, and Fei-Fei}]{imagenet}
Deng, J.; Dong, W.; Socher, R.; Li, L.-J.; Li, K.; and Fei-Fei, L. 2009.
\newblock Imagenet: A large-scale hierarchical image database.
\newblock In \emph{2009 IEEE conference on computer vision and pattern recognition}, 248--255. Ieee.

\bibitem[{Ding et~al.(2021)Ding, Zhang, Ma, Han, Ding, and Sun}]{ding2021repvgg}
Ding, X.; Zhang, X.; Ma, N.; Han, J.; Ding, G.; and Sun, J. 2021.
\newblock Repvgg: Making vgg-style convnets great again.
\newblock In \emph{Proceedings of the IEEE/CVF conference on computer vision and pattern recognition}, 13733--13742.

\bibitem[{Duda.(2013)}]{ANS:2013}
Duda., J. 2013.
\newblock {Asymmetric numeral systems: entropy coding combining speed of Huffman coding with compression rate of arithmetic coding}.
\newblock ArXiv:1311.2540.

\bibitem[{Goldman et~al.(2024)Goldman, Caciularu, Eyal, Cao, Szpektor, and Tsarfaty}]{unpackingtokenizationevaluatingtext}
Goldman, O.; Caciularu, A.; Eyal, M.; Cao, K.; Szpektor, I.; and Tsarfaty, R. 2024.
\newblock Unpacking Tokenization: Evaluating Text Compression and its Correlation with Model Performance.
\newblock arXiv:2403.06265.

\bibitem[{Hochreiter and Schmidhuber(1997)}]{lstm:1997}
Hochreiter, S.; and Schmidhuber, J. 1997.
\newblock Long short-term memory. Neural Computation.
\newblock \emph{Neural Computation}.

\bibitem[{Hoffmann et~al.(2022)Hoffmann, Borgeaud, Mensch, Buchatskaya, Cai, Rutherford, Casas, Hendricks, Welbl, Clark et~al.}]{chinchilla}
Hoffmann, J.; Borgeaud, S.; Mensch, A.; Buchatskaya, E.; Cai, T.; Rutherford, E.; Casas, D. d.~L.; Hendricks, L.~A.; Welbl, J.; Clark, A.; et~al. 2022.
\newblock Training compute-optimal large language models.
\newblock \emph{arXiv preprint arXiv:2203.15556}.

\bibitem[{Howard and Vitter(1991)}]{AC:1991}
Howard, P.~G.; and Vitter, J.~S. 1991.
\newblock {Analysis of arithmetic coding for data compression}.
\newblock In \emph{{Proceedings of Data Compression Conference (DCC)}}.

\bibitem[{Hu et~al.(2021)Hu, Shen, Wallis, Allen-Zhu, Li, Wang, Wang, and Chen}]{lora:2021}
Hu, E.~J.; Shen, Y.; Wallis, P.; Allen-Zhu, Z.; Li, Y.; Wang, S.; Wang, L.; and Chen, W. 2021.
\newblock LoRA: Low-Rank Adaptation of Large Language Models.
\newblock arXiv:2106.09685.

\bibitem[{Huang et~al.(2024)Huang, Zhang, Shan, and He}]{huang2024}
Huang, Y.; Zhang, J.; Shan, Z.; and He, J. 2024.
\newblock Compression Represents Intelligence Linearly.
\newblock arXiv:2404.09937.

\bibitem[{Huffman(1952)}]{Huffman:1952}
Huffman, D.~A. 1952.
\newblock {A method for the construction of minimum-redundancy codes}.
\newblock In \emph{{Proceedings of the IRE}}.

\bibitem[{HutterPrize(2006)}]{hutterprize}
HutterPrize. 2006.
\newblock {500'000€ Prize for Compressing Human Knowledge}.
\newblock \url{http://prize.hutter1.net/}.

\bibitem[{Kaplan et~al.(2020)Kaplan, McCandlish, Henighan, Brown, Chess, Child, Gray, Radford, Wu, and Amodei}]{scalinglaw}
Kaplan, J.; McCandlish, S.; Henighan, T.; Brown, T.~B.; Chess, B.; Child, R.; Gray, S.; Radford, A.; Wu, J.; and Amodei, D. 2020.
\newblock Scaling laws for neural language models.
\newblock \emph{arXiv preprint arXiv:2001.08361}.

\bibitem[{Knoll(2023)}]{cmix:2023}
Knoll, B. 2023.
\newblock {Cmix version 20, a lossless data compression program}.
\newblock \url{http://www.byronknoll.com/cmix.html}.

\bibitem[{Lester et~al.(2024)Lester, Lee, Alemi, Pennington, Roberts, Sohl-Dickstein, and Constant}]{trainingllmsneurallycompressed}
Lester, B.; Lee, J.; Alemi, A.; Pennington, J.; Roberts, A.; Sohl-Dickstein, J.; and Constant, N. 2024.
\newblock Training LLMs over Neurally Compressed Text.
\newblock arXiv:2404.03626.

\bibitem[{Loshchilov and Hutter(2019)}]{adamw}
Loshchilov, I.; and Hutter, F. 2019.
\newblock Decoupled Weight Decay Regularization.
\newblock arXiv:1711.05101.

\bibitem[{MacKay.(2003)}]{MacKay:2003}
MacKay., D. J.~C., ed. 2003.
\newblock \emph{Information theory, inference, and learning algorithms.}
\newblock Cambridge University Press.

\bibitem[{Mahoney(2006{\natexlab{a}})}]{paq8h}
Mahoney, M. 2006{\natexlab{a}}.
\newblock PAQ8H: A High Compression Ratio Data Compression Program.
\newblock \url{http://mattmahoney.net/dc/paq8h.zip}.
\newblock Accessed: 2024-08-09.

\bibitem[{Mahoney(2006{\natexlab{b}})}]{text-data}
Mahoney, M. 2006{\natexlab{b}}.
\newblock Text8 Dataset.
\newblock \url{http://mattmahoney.net/dc/textdata.html}.
\newblock Accessed: 2024-08-09.

\bibitem[{Mahoney(2024)}]{benchmark}
Mahoney, M. 2024.
\newblock {Large Text Compression Benchmark}.
\newblock \url{https://www.mattmahoney.net/dc/text.html}.

\bibitem[{Meta.(2015)}]{zstd}
Meta. 2015.
\newblock {ZSTD, Zstandard - Fast real-time compression algorithm}.
\newblock \url{https://github.com/facebook/zstd}.

\bibitem[{Panayotov et~al.(2015)Panayotov, Chen, Povey, and Khudanpur}]{librispeech}
Panayotov, V.; Chen, G.; Povey, D.; and Khudanpur, S. 2015.
\newblock Librispeech: an asr corpus based on public domain audio books.
\newblock In \emph{2015 IEEE international conference on acoustics, speech and signal processing (ICASSP)}, 5206--5210. IEEE.

\bibitem[{Pasco.(1977)}]{AC:1977}
Pasco., R.~C. 1977.
\newblock {Source coding algorithms for fast data compression (ph.d. thesis abstr).}
\newblock \emph{IEEE Trans. Inf. Theory}.

\bibitem[{Pasco.(1996)}]{gzip}
Pasco., R.~C. 1996.
\newblock {GZIP file format specification version 4.3.}
\newblock \emph{RFC 1952}.

\bibitem[{Peng et~al.(2023)Peng, Alcaide, Anthony et~al.}]{rwkv:2023}
Peng, B.; Alcaide, E.; Anthony, Q.; et~al. 2023.
\newblock RWKV: Reinventing RNNs for the Transformer Era.
\newblock arXiv:2305.13048.

\bibitem[{Schmidt et~al.(2024)Schmidt, Reddy, Zhang, Alameddine, Uzan, Pinter, and Tanner}]{tokenizationcompression}
Schmidt, C.~W.; Reddy, V.; Zhang, H.; Alameddine, A.; Uzan, O.; Pinter, Y.; and Tanner, C. 2024.
\newblock Tokenization Is More Than Compression.
\newblock arXiv:2402.18376.

\bibitem[{Sennrich, Haddow, and Birch(2016)}]{bpe}
Sennrich, R.; Haddow, B.; and Birch, A. 2016.
\newblock Neural Machine Translation of Rare Words with Subword Units.
\newblock arXiv:1508.07909.

\bibitem[{Seward(2000)}]{bzip2}
Seward, J. 2000.
\newblock {On the Performance of BWT Sorting Algorithms}.
\newblock \emph{Proceedings of the IEEE Data Compression Conference 2000}.

\bibitem[{Shannon.(1948)}]{shannon:1948}
Shannon., C.~E. 1948.
\newblock {A mathematical theory of communication.}
\newblock \emph{Bell Syst. Tech. J.}

\bibitem[{Touvron et~al.(2023{\natexlab{a}})Touvron, Lavril, Izacard, Martinet, Lachaux, Lacroix, Rozi{\`e}re, Goyal, Hambro, Azhar, Rodriguez, Joulin, Grave, and Lample}]{touvron2023llama}
Touvron, H.; Lavril, T.; Izacard, G.; Martinet, X.; Lachaux, M.-A.; Lacroix, T.; Rozi{\`e}re, B.; Goyal, N.; Hambro, E.; Azhar, F.; Rodriguez, A.; Joulin, A.; Grave, E.; and Lample, G. 2023{\natexlab{a}}.
\newblock LLaMA: Open and Efficient Foundation Language Models.
\newblock \emph{arXiv preprint arXiv:2302.13971}.

\bibitem[{Touvron et~al.(2023{\natexlab{b}})Touvron, Martin, Stone, Albert, Almahairi, Babaei, Bashlykov, Batra, Bhargava, Bhosale et~al.}]{llama2}
Touvron, H.; Martin, L.; Stone, K.; Albert, P.; Almahairi, A.; Babaei, Y.; Bashlykov, N.; Batra, S.; Bhargava, P.; Bhosale, S.; et~al. 2023{\natexlab{b}}.
\newblock Llama 2: Open foundation and fine-tuned chat models.
\newblock \emph{arXiv preprint arXiv:2307.09288}.

\bibitem[{Valmeekam et~al.(2023)Valmeekam, Narayanan, Kalathil, Chamberland, and Shakkottai}]{llmzip:2023}
Valmeekam, C. S.~K.; Narayanan, K.; Kalathil, D.; Chamberland, J.-F.; and Shakkottai, S. 2023.
\newblock LLMZip: Lossless Text Compression using Large Language Models.
\newblock arXiv:2306.04050.

\bibitem[{Vaswani et~al.(2017)Vaswani, Shazeer, Parmar, Uszkoreit, Jones, Gomez, Kaiser, and Polosukhin}]{transformer:2017}
Vaswani, A.; Shazeer, N.; Parmar, N.; Uszkoreit, J.; Jones, L.; Gomez, A.~N.; Kaiser, L.; and Polosukhin, I. 2017.
\newblock Attention Is All You Need.
\newblock arXiv:1706.03762.

\bibitem[{{Xiph.Org Foundation}(2001)}]{flac}
{Xiph.Org Foundation}. 2001.
\newblock Free Lossless Audio Codec.
\newblock \url{https://xiph.org/flac/}.
\newblock Accessed: 2024-08-09.

\bibitem[{Ziv(1977)}]{lz77:1977}
Ziv, A., Jacob;~Lempel. 1977.
\newblock {A Universal Algorithm for Sequential Data Compression}.
\newblock \emph{IEEE Transactions on Information Theory}.

\end{thebibliography}

\clearpage

\section*{Supplementary Material}

This supplementary material provides more experimental results of our paper (ID: 9661), which were omitted from the main paper due to space constraints. It includes detailed configurations of model structures, extended analysis of compression performance, and further discussion on decoding speeds, offering a comprehensive view of our L3TC method.

\paragraph{Model Structures across Different Sizes.}

For the proposed L3TC models, we achieve different model sizes by adjusting the number of layers, embedding dimensions, and hidden sizes, following the parameter tuning approach of the original RWKV models, as shown in \cref{tab:structures}.

\begin{table}[h]
    \centering
    \small
    \setlength{\tabcolsep}{9pt}
    \begin{tabularx}{0.95\columnwidth}{cccc}
    \toprule
    \textbf{Model} & \textbf{Layer} & \textbf{Embed dim} & \textbf{Hidden size} \\
    \midrule
    L3TC-200K & 2 & 96 & 96 \\[1pt]
    L3TC-800K & 2 & 176 & 192 \\[1pt]
    L3TC-3.2M & 3 & 256 & 512 \\[1pt]
    L3TC-12M  & 4 & 384  & 1024 \\[1pt]
    \bottomrule
    \end{tabularx}
    \captionsetup{font=small}
    \vspace{-2mm}
    \caption{\small Model structures across different sizes using RWKV blocks.}
    \label{tab:structures}
\end{table}

Specifically, the number of layers is varied to control the model's depth, with options ranging from 2 to 4 layers. The embedding dimension, which influences the capacity of the model's attention mechanism, is scaled between 96 and 384. The hidden size, which determines the width of the feedforward layers, is adjusted accordingly, ranging from 96 to 1024. 
Besides, the model structures of transformer and transformer-XL are listed in \cref{tab:structures2} and \cref{tab:structures3}. The number of head is set as $8$ for both transformer and transformerXL. 
These configurations allow us to scale the proposed L3TC models from smaller, more lightweight versions to larger, more powerful versions.

% For the proposed L3TC models, we achieve different model sizes by adjusting the number of layers, attention embedding dimensions, and hidden sizes, following the same parameter tuning approach as the original RWKV models, as shown in \cref{tab:structures}.

\begin{table}[h]
    \centering
    \small
    \setlength{\tabcolsep}{7pt}
    \begin{tabularx}{0.95\columnwidth}{cccc}
    \toprule
    \textbf{Model} & \textbf{Layer} & \textbf{hidden dim} & \textbf{FFN dim} \\
    \midrule
    Transformer-200K & 2 & 64 & 256 \\[1pt]
    Transformer-800K & 2 & 128 & 512 \\[1pt]
    Transformer-169M & 12 & 768 & 3072 \\[1pt]
    % Transformer-200K & 2 & 96 & 96 \\[1pt]
    % Transformer-800K & 2 & 176 & 192 \\[1pt]
    % Transformer-169M & 12 & 768 & 3072 \\[1pt]
    \bottomrule
    \end{tabularx}
    \captionsetup{font=small}
    \vspace{-2mm}
    \caption{Model structures across different sizes using Transformer blocks.}
    \label{tab:structures2}
\end{table}

\begin{table}[h]
    \centering
    \small
    \setlength{\tabcolsep}{7pt}
    \begin{tabularx}{0.95\columnwidth}{cccc}
    \toprule
    \textbf{Model} & \textbf{Layer} & \textbf{hidden dim} & \textbf{FFN dim} \\
    \midrule
    % Transformer-200K & 2 & 64 & 256 \\[1pt]
    % Transformer-800K & 2 & 128 & 512 \\[1pt]
    % Transformer-169M & 12 & 768 & 3072 \\[1pt]
    TransformerXL-200K & 2 & 96 & 96 \\[1pt]
    TransformerXL-800K & 2 & 176 & 192 \\[1pt]
    TransformerXL-169M & 12 & 768 & 3072 \\[1pt]
    \bottomrule
    \end{tabularx}
    \captionsetup{font=small}
    \vspace{-2mm}
    \caption{Model structures across different sizes using TransformerXL blocks.}
    \label{tab:structures3}
\end{table}

% \section{Supplementary Experimental Analysis}

\paragraph{Ablation Study on Model Structures.}

As shown in \cref{fig:rwkv} of the main paper, we construct the L3TC models by incorporating high-rank reparameterization (HiRA) and a linear shortcut to the RWKV models. To validate the effectiveness of these two modules, we evaluate three configurations using a character-based tokenizer with a 128 vocabulary size: 1) the original RWKV structure, 2) the RWKV structure with a linear shortcut, 3) RWKV with both a linear shortcut and HiRA (i.e., the proposed L3TC model), as presented in \cref{tab:ablation}. Notably, all these experiments use models with 200K parameters. HiRA is configured with a single bypass branch and a rank set to four times the main branch's dimension, which has been proved as the optimal configuration in the main paper.

\begin{table}[H]
    \centering
    \small
    \setlength{\tabcolsep}{7pt}
    \begin{tabularx}{0.95\columnwidth}{lll}
    \toprule
    \textbf{Model} & \textbf{MACs} & \textbf{CR(\%)} \\
    \midrule
    RWKV-200K & 143.6K & 24.36 \\[1pt]
    RWKV-200K + Linear &  160.9K & 24.18 {\footnotesize{(\color{blue}-0.18}})\\[1pt]
    RWKV-200K + Linear + HiRA &  160.9K & 23.51 {\footnotesize{(\color{blue}-0.85}}) \\[1pt]
    \bottomrule
    \end{tabularx}
    \captionsetup{font=small}
    \vspace{-2mm}
    \caption{Ablation study on model structures.}
    \label{tab:ablation}
\end{table}

As illustrated in \cref{tab:ablation}, adding the linear shortcut and HiRA improves the compression performance by about 0.18\% and 0.67\%, respectively, with a combined improvement of 0.85\%. While the linear shortcut slightly increases MACs from 143.6K to 160.9K, HiRA incurs no additional computational cost. Therefore, we simultaneously utilize the linear shortcut and HiRA to enhance compression performance without compromising computational efficiency.

\paragraph{More Results on Compression Performance.}

Due to space limitations, the main paper excludes the experimental analysis of NNCP~\cite{nncpv2:2021}, CMIX~\cite{cmix:2023}. Their compression details are provided herein, as illustrated in \cref{tab:cr}. It can be observed that advanced compressors like CMIX and NNCP achieve impressive compression performance, with compression ratios of approximately 11\%, which is comparable to RWKV-1.5B model. However, these models typically involve excessive computational complexity. CMIX and NNCP require 7.2 and 2.8 days, respectively, to decode 1 GB of text (with decoding speeds in bytes per second)~\cite{benchmark}. The overly high computational and memory demands render these approaches impractical for real-time usage.

\begin{table}[H]
    \centering
    \small
    \setlength{\tabcolsep}{8pt}
    \begin{tabular}{lrr}
    \toprule
    \textbf{Compressor} & \textbf{CR(\%)} & \textbf{Decoding Speed}\\
    \midrule
    CMIX v20~\cite{cmix:2023}  & 10.99 &  4 KB/s \\
    NNCP v3.2~\cite{nncpv2:2021}  & 10.66 & 1.6 KB/s\\
    RWKV-1.5B  &10.89  & 20 KB/s \\
    % \midrule
     % & $\text{RWKV-169M}^{\textbf{*}}$ & RWKV-Pile-tokenizer-50K & 130.71M & 14.23 & 48.03\\[1pt]
     % 2048 & $\text{RWKV-430M}^{\textbf{*}}$ & RWKV-Pile-tokenizer-50K & 378.84M & 12.38 & 98.38\\[1pt]
     %     & $\text{RWKV-3B}^{\textbf{*}}$   & RWKV-Pile-tokenizer-50K & 2.86G   & 10.33 & 610.33\\[1pt]
         % & Transformer-169M   & character-based & 85.11K & 14.81 & 31.71\\[1pt]
         % & TransformerXL-169M & character-based & 92.26M & 14.81 & 31.71\\[1pt]
         % & $\text{LLMZip}^{\textbf{*}}({\text{zlib}})$~\cite{llmzip:2023}  & Llama-tiktoken-32K  & 6.61G  & 13.11 & 1413.11\\[1pt]
    \bottomrule
    \end{tabular}
    \captionsetup{font=small}
    \vspace{-2mm}
    \caption{\small More results on compression performance.}
    \label{tab:cr}
\end{table}

% Due to page limitation, we omit the compression ratio of some pretrained models and NNCP~\cite{nncpv2:2021}, CMIX~\cite{cmix:2023}. 

\paragraph{Ablation study on Decoding Speed.}

To evaluate the effect of our proposed methods on decoding speed, we evaluate the L3TC-200K models using a single HiRA bypass branch with its rank set to four times the dimension of the main branch. Two tokenizers are compared, i.e. the proposed outlier-aware tokenizer with a 16K vocabulary size and $0.999$ coverage, and the character-based tokenizer with a 128 vocabulary size. As shown in \cref{tab:dtime-bs128}, with a batch size of $128$ and automatically running on the Apple Neural Engine (ANE). 
When using the same (character-based) tokenizer, the proposed L3TC models and equivalently sized RWKV models exhibit nearly identical decoding speeds, confirming that the high-rank parameterization has minimal impact on inference complexity. 
However, when paired with our proposed outlier-aware tokenizer, our proposed method is $2.2\times$ faster than vanilla RWKV models, from 376 to 842 KB/s, owing to our proposed outlier-aware tokenizer. This improvement occurs because the outlier-aware tokenizer merges frequent subwords, enabling the processing of more characters in a single inference, whereas the character-based tokenizer processes only one character at a time, causing less efficient data handling.

\begin{table}[H]
    \centering
    \small
    \setlength{\tabcolsep}{6pt}
    \begin{tabular}{lll}
    \toprule
       \textbf{Compressor}  &  \textbf{Tokenizer}  &  \makecell{\textbf{Decoding Speed} \\[3pt] \footnotesize{\textbf{(iPhone12@ANE)}}}\\
       \midrule
       RWKV-200K   &  character-based  &  376 KB/s\\[1pt]
       L3TC-200K (+HiRA)   &  character-based  &  376 KB/s\\[1pt]
       L3TC-200K   &  outlier-aware    &  842 KB/s \footnotesize{{\color{blue}(2.2$\times$)}}\\[1pt]
       \bottomrule
    \end{tabular}
    \captionsetup{font=small}
    \vspace{-2mm}
    \caption{\small Decoding Speeds on iPhone12 (ANE) with batch size 128.}
    \label{tab:dtime-bs128}
\end{table}

\paragraph{More results on Devices.}

When running on an iPhone, we tested the decoding speeds using both the ANE and GPU as backends. The runtime of the RWKV and L3TC models on an iPhone 12 is shown in \cref{tab:dtime-bs1}. When the batch size is reduced to 1, as indicated in \cref{tab:dtime-bs1}, the models automatically switch to the CPU, resulting in significantly slower decoding speeds, ranging from 11 to 27 KB/s. However, even under these conditions, our proposed outlier-aware tokenization method still demonstrates a speed improvement, achieving approximately a $2.5\times$ increase in decoding speed.
Besides, the high-rank reparameteration method does not bring any complexity overhead.

\begin{table}[H]
    \centering
    \small
    \setlength{\tabcolsep}{6pt}
    \begin{tabular}{lll}
    \toprule
       \textbf{Compressor}  &  \textbf{Tokenizer}  &  \makecell{\textbf{Decoding Speed} \\[3pt] \footnotesize{\textbf{(iPhone12@CPU)}}}\\
       \midrule
       RWKV-200K   &  character-based  &  11 KB/s\\[1pt]
       L3TC-200K (+HiRA)   &  character-based  &  11 KB/s\\[1pt]
       L3TC-200K   &  outlier-aware    &  27 KB/s \footnotesize{{\color{blue}(2.5$\times$)}}\\[1pt]
       \bottomrule
    \end{tabular}
    \captionsetup{font=small}
    \vspace{-2mm}
    \caption{\small Decoding Speeds on iPhone12 (CPU) with batch size 1.}
    \label{tab:dtime-bs1}
\end{table}

\end{document}